\pdfoutput=1

\documentclass[11pt]{article}

\usepackage[]{ACL2023}
\usepackage{xltabular}
\usepackage{times}
\usepackage{latexsym}
\usepackage{amsmath}
\usepackage{amssymb}

\usepackage[T1]{fontenc}

\usepackage[utf8]{inputenc}

\usepackage{microtype}

\usepackage{inconsolata}

\usepackage{algpseudocode}
\usepackage{algorithm}
\usepackage{hyperref}

\usepackage{caption}

\title{Diversity-Aware Coherence Loss for Improving Neural Topic Models}

\author{Raymond Li\footnotemark[2]~, Felipe González-Pizarro\footnotemark[2]~, Linzi Xing\footnotemark[2]~, Gabriel Murray\footnotemark[3]~, Giuseppe Carenini\footnotemark[2]\\
\footnotemark[2]\hspace{.2em} University of British Columbia, Vancouver, BC, Canada \\
\footnotemark[3]\hspace{.2em} University of Fraser Valley, Abbotsford, BC, Canada \\
\texttt{\{\href{mailto:raymondl@cs.ubc.ca}{\texttt{raymondl}}, \href{mailto:felipegp@cs.ubc.ca}{\texttt{felipegp}}, \href{mailto:lzxing@cs.ubc.ca}{\texttt{lzxing}}, \href{mailto:carenini@cs.ubc.ca}{\texttt{carenini}}\}@cs.ubc.ca} \\
\href{mailto:gabriel.murray@ufv.ca}{\texttt{gabriel.murray@ufv.ca}}
}

\usepackage{graphics}
\usepackage{arydshln}
\usepackage[normalem]{ulem}

\usepackage{color,colortbl,soulutf8}
\usepackage{float}

\definecolor{red}{RGB}{252, 243, 207}
\definecolor{red}{RGB}{255,204,204}
\definecolor{white}{RGB}{255,255,255}

\usepackage{booktabs}

\usepackage{ifthen}
\newif\ifshort
\shorttrue   
\ifshort
	\newcommand{\isShort}{true}
\else
	\newcommand{\isShort}{false}
\fi
\newcommand{\shortVer}[1]{\ifthenelse{\equal{\isShort}{true}}{{#1}}{}}
\newcommand{\longVer}[1]{\ifthenelse{\equal{\isShort}{false}}{{#1}}{}}

\ifshort
\usepackage[]{flushend}
\else
\fi

\newif\ifcomment
\commenttrue
\ifcomment
\newcommand{\fg}[1]{{\bf\textcolor{purple}{FG: #1}}}
\newcommand{\rl}[1]{{\bf\textcolor{blue}{RL: #1}}}
\newcommand{\lx}[1]{{\bf\textcolor{violet}{LX: #1}}}

\else
\newcommand{\fg}[1]{}
\newcommand{\rl}[1]{}
\newcommand{\lx}[1]{}

\fi

\newcommand\change[1]{{\leavevmode\color{red}#1}}
\newcommand\cut[1]{\textcolor{gray}{\sout{#1}}}

\renewcommand{\change}[1]{{\leavevmode\color{black}#1}} 
\renewcommand{\cut}[1]{\unskip}

\usepackage{comment}
\usepackage{tabularx,booktabs}
\usepackage{longtable}
\newcolumntype{s}{>{\hsize=.3\hsize}X}

\begin{document}
\maketitle
\begin{abstract}
The standard approach for neural topic modeling uses a variational autoencoder (VAE) framework that jointly minimizes the KL divergence between the estimated posterior and prior, in addition to the reconstruction loss. Since neural topic models are trained by recreating individual input documents, they do not explicitly capture the coherence between \change{topic} words on the corpus level. In this work, we propose a novel diversity-aware coherence loss that encourages the model to learn corpus-level coherence scores while maintaining a high diversity between topics. Experimental results on multiple datasets show that our method significantly improves the performance of neural topic models without requiring any  pretraining or additional parameters.
\end{abstract}

\section{Introduction}
The main goal of topic modeling is to discover latent topics that best explain the observed documents in the corpus.
The topics, conceptualized as a multidimensional distribution over the vocabulary, are useful
for many
downstream applications, including summarization 
\citep{wang-etal-2020-friendly, xiao2022attend},
text generation \citep{wang-etal-2019-topic, nevezhin-etal-2020-topic}, dialogue modeling \citep{xu2021topic, zhu-etal-2021-topic}\change{, as well as analyzing the data used for pretraining large language models \citep{chowdhery2022palm}}.
When presented to humans,
they are often represented as lists of the most 
probable words to assist the users in exploring and understanding the underlying themes in a large collection of documents.
While the extrinsic quality of topics can be quantified by the performance of their downstream \change{tasks}, 
the intrinsic interpretability of topics 
appears to be strongly correlated with
two important factors, namely \textit{coherence} and \textit{diversity} \citep{dieng-etal-2020-topic}.

The topic \textit{coherence} measures to what extent the words within a topic are related to each other in a meaningful way. Although human studies provide a direct method for evaluation, they can be costly,
especially when a large number of models
are waiting to be assessed.
Therefore, various automatic metrics have been developed to measure topic coherence
\citep{newman-etal-2010-automatic, mimno-etal-2011-optimizing, xing-etal-2019-evaluating, terragni2021word}. 
For instance, the 
well-established Normalized Pointwise Mutual Information (NPMI) metric \citep{lau-etal-2014-machine},
based on word co-occurrence 
within a fixed window, has been found to have a strong correlation with human judgment 
\citep{roder2015exploring}. On the other hand, topic \textit{diversity} measures 
to what extent the topics are able to capture different aspects of the corpus based on the uniqueness of the topic words \citep{nan-etal-2019-topic}. 
Importantly, studies have shown that optimizing for coherence can come at the expense of diversity \citep{burkhardt2019decoupling}.
Even without accounting for topic diversity, directly optimizing for topic coherence by itself is a non-trivial task, due to the computational overhead and non-differentiability of the score matrix~\citep{ding-etal-2018-coherence}.


While traditional topic modeling 
 algorithms are in the form of statistical models such as the Latent Dirichlet Allocation (LDA) \citep{blei2003latent}, advancements in variational inference methods 
\citep{kingma2013auto, rezende2014stochastic} 
have led to the rapid development of neural topic model (NTM) architectures \citep{miao2016neural, miao2017discovering, srivastava2017autoencoding}. More recently, follow-up works have focused on the integration of additional knowledge to improve the coherence of NTMs. Their attempts include the incorporation of external embeddings \citep{ding-etal-2018-coherence, card-etal-2018-neural, dieng-etal-2020-topic, bianchi-etal-2021-pre, bianchi-etal-2021-cross}, knowledge distillation \citep{hoyle-etal-2020-improving}, and model pretraining \citep{zhang-etal-2022-pre}.
However, as the model is designed to operate on a document-level input, one significant limitation 
of NTMs is their inability to explicitly capture \cut{coherence on the corpus level} \change{the corpus-level coherence score, which assesses 
the extent to which words within specific topics tend to occur together in a comparable context within a given corpus.
For example, semantically irrelevant words such as ``\textit{politics}'' and ``\textit{sports}'' might be contextually relevant in a given corpus (e.g., government funding for the national sports body).
}
Recently, one closely related work addresses this gap by reinterpreting topic modeling as a coherence optimization task with diversity as a constraint \citep{lim-lauw-2022-towards}. 

While traditional topic models tend to directly use
corpus-level coherence signals,
such as factorizing the document-term 
matrix \citep{steyvers2007probabilistic}, and topic segment labeling with random walks on co-occurrence graphs \citep{mihalcea2011graph, joty2013topic}, to the best of our knowledge, no existing work have explicitly integrated corpus-level coherence \change{scores} into the training of NTMs without sacrificing topic diversity. To address this gap, we propose a novel \textbf{coherence-aware diversity loss}, which is effective to improve both the coherence and diversity of NTMs by adding as an auxiliary loss during training. Experimental results show that this method can significantly improve baseline models without any pretraining or additional parameters\footnote{The implementation of our work is available at: \href{https://github.com/raymondzmc/Topic-Model-Diversity-Aware-Coherence-Loss}{https://github.com/raymondzmc/Topic-Model-Diversity-Aware-Coherence-Loss}}. 





\section{Background}


Latent Dirichlet Allocation (LDA)~\citep{blei2003latent}
is a simple yet effective probabilistic generative model trained on a
collection of documents.
It is based on the assumption that each document $w$ in the corpus is described by a random mixture of latent topics $z$ sampled from a distribution parameterized by $\theta$, where the topics $\beta$ are represented as a multidimensional distribution over the vocabulary $V$.
The formal algorithm describing the generative process is presented in \autoref{sec:lda}.
Under this assumption, the marginal likelihood of the document $p(w|\alpha, \beta)$ is described as: 
\begin{align}
    \int_\theta \Bigg( \prod_{i}^{|V|} \sum_{z_i}^{K}p(w_i|z_i, \beta)p(z_i|\theta)\Bigg )p(\theta|\alpha)d\theta
\label{eq:lda-marginal-likelihood}
\end{align}

However, since the posterior distribution $p(z_i|\theta)$ is intractable for exact inference, a wide variety of approximate inference algorithms have been used for LDA (e.g., \citet{hoffman2010online}).

A common strategy to approximate 
such posterior is employing the variational auto-encoder (VAE) \citep{kingma2013auto}.
In particular, NTMs use
an encoder network to compress the document representation into a continuous latent distribution and pass it to a generative decoder to reconstruct the bag-of-words (BoW) representation of the documents.
The model is trained to minimize the evidence lower bound (ELBO) of the marginal log-likelihood described by the LDA generative process:

\begin{align}
\begin{split}
    L_{\textrm{ELBO}} = &- D_{\textrm{KL}}[q(\theta, z|w) || p(\theta, z|\alpha)] \\
    &+\mathbb{E}_{q(\theta, z|w)}[\log p(w|z, \theta, \alpha, \beta)]
\end{split}
\label{eq:elbo}
\end{align}

In \autoref{eq:elbo}, the first term attempts to match the variational posterior over latent variables
to the prior,
and the second term ensures that the variational posterior favors
values of the latent variables that are good at explaining the data (i.e., reconstruction loss). While standard Gaussian prior has typically been used in VAEs,
\textbf{ProdLDA}~\citep{srivastava2017autoencoding} showed that 
using a Laplace approximation of the Dirichlet prior 
achieved superior performance. 
To further improve topic coherence, \textbf{CombinedTM}~\citep{bianchi-etal-2021-pre} concatenated the BoW input with contextualized SBERT embeddings \citep{reimers-gurevych-2019-sentence}, while \textbf{ZeroshotTM}~\citep{bianchi-etal-2021-cross} used only contextualized embeddings as input. These are the three baselines included in our experiments.
%

\section{Proposed Methodology}
Despite the recent advancements, one significant limitation
of the NTM is that since the model is trained on document-level input, it does not have direct access to corpus-level coherence information (i.e., word co-occurrence).
Specifically,
the topic-word distribution $\beta$ is optimized on the document-level reconstruction loss, which may not be an accurate estimate of the true corpus distribution due to the inherent stochasticity of gradient-descent algorithms.
We address this problem by explicitly integrating a corpus-level coherence metric into the training process of NTMs using an auxiliary loss. 

\subsection{Optimizing Corpus Coherence}
To improve
the topic-word distribution $\beta$, we
maximize the corpus-level coherence through the well-established NPMI metric\footnote{Detailed definition of NPMI is presented in \autoref{sec:npmi}.} \citep{bouma2009normalized, lau-etal-2014-machine}.
After computing the pairwise NPMI matrix $N \in \mathbb{R}^{|V|\times |V|}$ on the corpus, 
we use the negative $\beta$-weighted NPMI scores of the top-$n$ words within each topic as the
weight for the coherence penalty of $\beta$, where $n$ is a hyperparameter that equals to the number of topic words to use. Specifically, we apply a mask $M_c$ to keep the top-$n$ words of each topic and apply the row-wise softmax operation $\sigma$ to ensure the value of the penalty is always positive. We define the coherence weight $W_C$ in \autoref{eq:coherence-loss}.

\begin{equation}
    W_C = 1 - \textrm{normalize}(\sigma(\beta \odot M_c)N)
    \label{eq:coherence-loss}
\end{equation}

Intuitively, each value in $\sigma(\beta \odot M_{k})N$ represents the $\beta$-weighted average NPMI score with other words in the topic. Then we use row-wise normalization to scale the values, so $W_C \in [0, 1]$.


\subsection{Improving Topic Diversity}
One problem with the coherence weight $W_C$ is that it does not consider the diversity across topics. 
To account for this, we propose an additional method to
simultaneously 
improve topic diversity by encouraging words unused by other topics to have higher probabilities. To achieve this, we bin the words within each topic into two groups, where the words in the first group consist of those that already have a high probability in other topics (i.e., appear within top-$n$ words), while the second group does not. The intuition is that we want to penalize the words in the first group more than the words in the second group. In practice, we 
use a mask $M_d \in \mathbb{R}^{K \times V}$ for selecting $\beta$ logits in the first group, where hyperparameter $\lambda_d \in [0.5, 1]$ is a balancing constant between the two groups
and $n$ is the number of topic words to use. We then compute the diversity-aware coherence weight $W_D$ as the $\lambda_d$-weighted sum of $W_C$:

\begin{equation}
    W_D = \lambda_d M_d \odot W_C + (1 - \lambda_d) (\neg M_d) \odot W_C
\label{eq:weighted-w}
\end{equation}

From \autoref{eq:weighted-w}, we see that when $\lambda_d = 0.5$, there are no constraints on diversity since the two groups are penalized equally ($2W_D = W_C$). 

\subsection{Auxiliary Loss}

From the two definitions of coherence weight ($W_C, W_D$), we propose an auxiliary loss that can be directly combined with the ELBO loss (\autoref{eq:elbo}) when training the NTM. 
Since $\beta$ are unnormalized logits containing negative values, we apply the softmax operation $\sigma(\beta)$ 
to avoid unbound optimization.

\begin{equation}
    L_{\textrm{AUX}} = \frac{1}{2}[\sigma(\beta)]^2 \odot W_D
\label{eq:aux-loss}
\end{equation}

In \autoref{eq:aux-loss}, the topic probabilities are penalized by their negative weighted coherence score with the top-$n$ words. The square operation ensures that words with very high probability are penalized to avoid the global minima, we justify this decision based on its partial derivatives in the next subsection. 

The final objective function is the multitask loss consisting of the ELBO and our defined auxiliary loss:
\begin{equation}
    L = L_{\textrm{ELBO}} + \lambda_a L_{\textrm{AUX}}
\end{equation}

During training, we employ a linear warm-up schedule to increase $\lambda_a$ gradually, so the model can learn to reconstruct the BoW representation based on the topic distribution $\alpha$ before optimizing for coherence and diversity. 

\subsection{Derivatives}
We justify our auxiliary loss defined in \autoref{eq:aux-loss} using the derivatives w.r.t. the $\beta$ parameters. For simplicity, we define $p_{k, i}=\sigma(\beta_k)_i$ as the softmax probability for word $i$ in topic $k$. Since we detach the gradients when computing $W$, it can be treated as a constant $w$ in the derivatives.
\begin{align}
\begin{split}
    \frac{\partial L_{\textrm{AUX}}}{\partial \beta_{k, i}} =~& w \cdot p_{k, i}\cdot p_{k, i}(1 - p_{k, i}) ~+ \\
    & w \cdot \sum_{j\neq i} p_{k, j} (-p_{k, j}p_{k, i})
\end{split}
\label{eq:gradients}
\end{align}

In \autoref{eq:gradients}, the partial derivatives w.r.t. $\beta_{k,i}$ can be broken down into two terms. In the first term, the softmax derivative $p_{k, i}(1 - p_{k, i})$ is zero when $p_{k, i}$ is either 0 or 1 (really small or really large). The additional $p_{k, i}$ (from the square operation) penalizes over-confident logits and leads to better topics. Similarly for the second term, since $\sum_i p_{k, i} = 1$, $\sum_{j\neq i}\big(p_{k, j}p_{k, i}\big)$ is zero (global minima) when one logit dominates the others. Therefore, the additional $p_{k, j}$ has the same penalizing effect on the over-confident logits.

\begin{table*}[ht]
\centering
\resizebox{\textwidth}{!}{%
\footnotesize
\begin{tabular}{lrrrr|rrrr|rrrr}
\toprule
\multicolumn{1}{c}{Dataset} &
  \multicolumn{4}{c}{20NewsGroup} &
  \multicolumn{4}{c}{Wiki20K} &
  \multicolumn{4}{c}{GoogleNews} \\
  \midrule
\multicolumn{1}{c}{Metrics} &
  \multicolumn{1}{|c}{NPMI} &
  \multicolumn{1}{c}{WE} &
  \multicolumn{1}{c}{I-RBO} &
  \multicolumn{1}{c}{TU} &
  \multicolumn{1}{|c}{NPMI} &
  \multicolumn{1}{c}{WE} &
  \multicolumn{1}{c}{I-RBO} &
  \multicolumn{1}{c}{TU} &
  \multicolumn{1}{|c}{NPMI} &
  \multicolumn{1}{c}{WE} &
  \multicolumn{1}{c}{I-RBO} &
  \multicolumn{1}{c}{TU} \\ 
  \midrule
\multicolumn{1}{l|}{LDA} &
  .0426 &
  .1624 &
  \cellcolor{red!55}\textbf{.9880} &
  \multicolumn{1}{l|}{\cellcolor{red!55}\textbf{.8077}} &
  -.0470 &
  .1329 &
  .9934 &
  \multicolumn{1}{l|}{\cellcolor{red!55}\textbf{.8664}} &
  -.2030 &
  .0989 &
  .9973 &
  \cellcolor{red!55}\textbf{.9065} \\
\multicolumn{1}{l|}{ProdLDA} &
  .0730 &
  .1626 &
  .9923 &
  \multicolumn{1}{l|}{.7739} &
  .1712 &
  .1883 &
  \cellcolor{red!55}\textbf{.9948} &
  \multicolumn{1}{l|}{.7674} &
  .0919 &
  .1240 &
  .9974 &
  .8460 \\
\multicolumn{1}{l|}{CombinedTM} &
  .0855 &
  .1643 &
  .9922 &
  \multicolumn{1}{l|}{.7705} &
  .1764 &
  .1893 &
  .9941 &
  \multicolumn{1}{l|}{.7509} &
  .1062 &
  .1316 &
  .9943 &
  .7498 \\
\multicolumn{1}{l|}{ZeroshotTM} &
  .1008 &
  .1749 &
  .9910 &
  \multicolumn{1}{l|}{.7214} &
  .1783 &
  .1896 &
  .9916 &
  \multicolumn{1}{l|}{.6999} &
  .1218 &
  .1321 &
  .9967 &
  .8200 \\ 
  \midrule 

\multicolumn{1}{l|}{ProdLDA + $W_C$} &
  .1233 &
  .1775 &
  .9916 &
  \multicolumn{1}{l|}{.7526} &
  .2386 &
  .2094 &
  .9905 &
  \multicolumn{1}{l|}{.6933} &
  .1236 &
  .1262 &
  .9973 &
  .8400 \\
\multicolumn{1}{l|}{CombinedTM + $W_C$} &
  .1301 &
  .1781 &
  .9910 &
  \multicolumn{1}{l|}{.7477} &
  .2392 &
  .2113 &
  .9890 &
  \multicolumn{1}{l|}{.6748} &
  .1378 &
  .1339 &
  .9938 &
  .7421 \\
\multicolumn{1}{l|}{ZeroshotTM + $W_C$} &
  .1456 &
  .1882 &
  .9895 &
  \multicolumn{1}{l|}{.6975} &
  .2455 &
  .2147 &
  .9862 &
  \multicolumn{1}{l|}{.6350} &
  .1562 &
  .1349 &
  .9964 &
  .8131 \\ 
  \midrule
\multicolumn{1}{l|}{ProdLDA + $W_D$} &
  .1235 &
  .1786 &
  .9940 &
  \multicolumn{1}{l|}{.7901} &
  .2367 &
  .2101 &
  .9929 &
  \multicolumn{1}{l|}{.7556} &
  .1275 &
  .1274 &
  \cellcolor{red!55}\textbf{.9975} &
  .8504 \\
\multicolumn{1}{l|}{CombinedTM + $W_D$} &
  .1309 &
  .1790 &
  .9935 &
  \multicolumn{1}{l|}{.7833} &
  .2404 &
  .2137 &
  .9918 &
  \multicolumn{1}{l|}{.7366} &
  .1429 &
  .1354 &
  .9942 &
  .7541 \\
\multicolumn{1}{l|}{ZeroshotTM + $W_D$} &
  \cellcolor{red!55}\textbf{.1482} &
  \cellcolor{red!55}\textbf{.1899} &
  .9919 &
  \multicolumn{1}{l|}{.7343} &
  \cellcolor{red!55}\textbf{.2460} &
  \cellcolor{red!55}\textbf{.2156} &
  .9890 &
  \multicolumn{1}{l|}{.6904} &
  \cellcolor{red!55}\textbf{.1569} &
  \cellcolor{red!55}\textbf{.1350} &
  .9967 &
  .8228 \\ 
  
  \bottomrule
\end{tabular}
}
\caption{Average results over $5$ number of topics ($K = 25, 50, 75, 100, 150$), where the results for each $K$ are averaged over $10$ random seeds. The results are reported for $\lambda_d = 0.7$, a mid-range value in the $[0.5, 1]$ interval.}
\label{tab:main-results}
\end{table*}

\section{Experiments}
In this section, we describe the experimental settings and present the quantitative results to assess the benefits of our proposed loss.

\subsection{Datasets and Evaluation Metrics}
To test the 
generality of our approach, we train and evaluate our models on three publicly available datasets: 20NewsGroups, Wiki20K~\citep{bianchi-etal-2021-cross}, and GoogleNews~\citep{qiang2020short}. 
We provide the statistics of the three datasets in 
\autoref{tab:datasets_stats}\footnote{Detailed description of the three datasets is provided in \autoref{sec:datasets}.}.
\begin{table}[ht]
\centering
\begin{tabular}{@{}lrrr@{}}
\toprule
Dataset & Domain & Docs  & Vocabulary  \\
\midrule
20Newsgroups & Email & 18,173 & 2,000    \\
 Wiki20K & Article & 20,000 & 2,000     \\
 Google News & News & 11,108 &8,110     \\
 \bottomrule
\end{tabular}
\caption{Statistics of the three datasets used in our experiments.} 
\label{tab:datasets_stats}
\end{table}


We use automatic evaluation metrics to measure  the topic coherence and diversity of the models. For coherence, we use the NPMI and Word Embedding (WE)~\cite{fang2016using} metrics, which measure the pairwise NPMI score and word embedding similarity, respectively,  between the top-$10$ words of each topic. For diversity, we use Topic Uniqueness (TU)~\cite{dieng-etal-2020-topic}, which measures the proportion of unique topic words, and Inversed Rank-Biased Overlap (I-RBO) \cite{terragni2021word,bianchi-etal-2021-pre}, 
measuring the rank-aware difference between all combinations of topic pairs.


\subsection{Baselines}
We plug our proposed auxiliary loss to three baseline NTMs' training process to demonstrate the benefits of our approach across different settings. Specifically, the three models are (1) ProdLDA \citep{srivastava2017autoencoding}, (2) CombinedTM \citep{bianchi-etal-2021-pre}, and (3) ZeroshotTM \citep{bianchi-etal-2021-cross}. \change{For comparison, we also include the results of the standard LDA algorithm \citep{blei2003latent}.}
\cut{We train all our models with 100 epochs and a learning rate of 0.002 following the settings reported by~\mbox{\citep{bianchi-etal-2021-pre}}}

\subsection{Hyperparemeter Settings}
\change{We follow the training settings reported by \citet{bianchi-etal-2021-pre}, with 100 epochs and a batch size of 100. The models are optimized using the ADAM optimizer \citep{kingma2014adam} with the momentum set to 0.99 and a fixed learning rate of $0.002$. We do not modify the architecture of the models, where the inference network is composed of a single hidden layer and 100 dimensions of softplus activation units~\citep{zheng2015improving}. The priors over the topic and document distributions are learnable parameters. A 20\% Dropout \citep{srivastava2014dropout} is applied to the document representations. During our evaluation, we follow the same setup and used the top-$10$ words of each topic for the coherence and diversity metrics.}

\change{For the hyperparameters introduced in the diversity-aware coherence loss, both $M_c$ and $M_d$ are computed using the top-$\mathbf{20}$ words of each topic. The scaling factor $\lambda_a$ is linearly increased for the first $50$ epochs and kept constant for the last $50$ epochs, we set $\lambda_a$ to be $\mathbf{100}$ in order to balance the loss magnitude of $L_{\textrm{ELBO}}$ and  $L_{\textrm{AUX}}$. The $\lambda_d$ in the diversity loss is set by taking a mid-range value of $\mathbf{0.7}$ in the $[0.5, 1]$ range. We do not perform any searches over our defined hyperparameters; we believe that additional experiments will yield better results (i.e., by using a validation set).}

\subsection{Results}
\autoref{tab:main-results} shows
improvements across all settings. 
However, with 
the basic coherence loss ($W_C$), 
the significant coherence increase comes at the expense of topic diversity, where a slight decrease can be observed in the I-RBO and TU scores.
In contrast, 
with the diversity-aware coherence loss ($W_D$), we observe that the model 
improves in coherence while having a significantly higher diversity over the basic loss ($W_C$).
The further coherence improvements can be attributed to the regularization effects, where words with a high probability of belonging to another topic are less likely to be related to words in the current topic. \change{Lastly, it is worth noting that due to the gradual increase in $\lambda_a$, our proposed loss has a negligible effect on the original document-topic distribution $\theta$, and only modifies the word distribution within the established topics. We provide some sample model outputs in \autoref{sec:model-output}.}

\subsection{Coherence and Diversity Trade-off}
To study the effects of $\lambda_d$ on the trade-off between coherence and diversity, we perform experiments with different values of $\lambda_d$ with the ZeroshotTM baseline, which has the best overall performance. 
Note that when $\lambda_d = 0.5$, the objective is equivalent to the 
basic coherence loss. From results on the 20NewsGroups Dataset (\autoref{tab:lambda-trade-off}), we see that coherence peaks at $\lambda_d=0.7$ before the diversity penalty begins to dominate the loss. Further, while a higher value of $\lambda_d$ leads to a lower coherency score, both coherency and diversity are still improved over the baselines for all values of $\lambda_d$, demonstrating the effectiveness of our method without the need for extensive hyperparameter tuning. We observe an identical trend in other datasets.

\begin{table}[ht]
\centering
\small
\begin{tabular}{@{}lrrrr@{}}
\toprule
 & NPMI & WE & I-RBO & TU  \\
\midrule
ZeroshotTM & .1008 &.1749 & .9910 & .7214  \\
$\lambda_d = 0.5$ & .1456 & .1882 & .9895 & .6975  \\
$\lambda_d = 0.6$ & .1428 & .1875 & .9908 & .7198  \\
$\lambda_d = 0.7$ & \cellcolor{red!55}\textbf{.1482}	& \cellcolor{red!55}\textbf{.1899} & .9919 &.7343 \\
$\lambda_d = 0.8$  & .1443 & .1890 & .9925 &.7499  \\
$\lambda_d = 0.9$  & .1369	& .1867 & .9933 &.7724  \\
$\lambda_d = 1.0$ & .1193 & .1816 & \cellcolor{red!55}\textbf{.9951} & \cellcolor{red!55}\textbf{.8086}  \\
 \bottomrule
\end{tabular}
\caption{Results on the 20NewsGroups dataset for different values of $\lambda_d$ with ZeroshotTM.} 
\label{tab:lambda-trade-off}
\end{table}
\vspace{-1em}

\subsection{Comparison with Composite Activation}

\change{The recent work by \citet{lim-lauw-2022-towards} proposed a model-free technique to refine topics based on the parameters of the trained model. Specifically, they solve
an optimization problem (with the NPMI score as the objective) using a pool of candidates while setting the diversity score as a constraint. 
}

\change{
Since their goal is similar to ours, we run further evaluations to compare the respective approaches. 
In particular,
we experiment with ZeroshotTM on the 20NewsGroups dataset for $K = 25, 50$. For comparison, we use their Multi-Dimensional Knapsack Problem (MDKP) formulation, 
since it achieved the best overall  performance. 
Regrettably, considering larger topic numbers was not possible due to the NP-hard runtime complexity of MDKP. From the results in \autoref{tab:mdkp-comparsion}, we see that while our methods have similar coherence scores, MDKP archives higher topic diversity due to its selectivity of less-redundant topics. However, when combining MDKP with our proposed loss (+ $W_D$ + MDKP), we achieve the highest overall performance across all metrics. This is expected since the pool of potential topic candidates is generated based on the trained model, and better-performing models lead to superior candidates.}

\begin{table}[ht]
\centering
\small
\begin{tabular}{@{}lrrrr@{}}
\toprule
 $K = 25$ & NPMI & WE & I-RBO & TU \\
\midrule
ZeroshotTM & .1059 & .1791 & .9927 & .9152  \\
+ MDKP & .1481 & .1895 & \cellcolor{red!55}\textbf{.9991} & .9804 \\ 
+ $W_D$ & .1433 & .1921 & .9981 & .9688 \\
+ $W_D$ + MDKP & \cellcolor{red!55}\textbf{.1657} & \cellcolor{red!55}\textbf{.2043} &.9989 & \cellcolor{red!55}\textbf{.9808}  \\
\bottomrule
\toprule
 $K = 50$ & NPMI & WE & I-RBO & TU \\
\midrule
ZeroshotTM & .1109 & .1746 & .9937 & .8498  \\
+ MDKP & .1578 & .1903 & .9983 & .9452 \\ 
+ $W_D$ & .1581 & .1921 & .9963 & .8840 \\
+ $W_D$ + MDKP & \cellcolor{red!55}\textbf{.1783} & \cellcolor{red!55}\textbf{.1932} & \cellcolor{red!55}\textbf{.9985} & \cellcolor{red!55}\textbf{.9500}  \\
\bottomrule
\end{tabular}
\caption{Results for comparing our approach with Composite Activation on the 20NewsGroups dataset. 
} 
\label{tab:mdkp-comparsion}
\end{table}
\vspace{-.8em}

\section{Conclusion and Future Work}
In this work, we present a novel diversity-aware coherence loss to simultaneously improve the coherence and diversity of neural topic models. 
In contrast to previous methods, our approach directly integrates corpus-level coherence scores into the training of Neural Topic Models.
The extensive experiments show that our proposal significantly improves the performance across all settings without requiring any pretraining or additional parameters. 

\change{For future work, we plan
to perform extensive user studies to examine the extent to which improvements in quantitative metrics affect human preference. Further, we would like to extend our approach to other quantitative metrics (e.g., semantic similarity), and perform extrinsic evaluation to study the effects 
of our approach 
when the topics are used for downstream tasks (e.g., summarization, dialogue modeling, text generation).}
\cut{Concurrent work by  has very recently proposed a model-free technique to refine  topics by solving an optimization problem (with the NPMI score as the objective) using a pool of potential candidates while setting the diversity score as a constraint. Since their goals are similar to ours, we plan to compare our respective approaches, especially exploring if they can be synergistically combined. Something we expect to work rather well, because our model is designed to improve the quality of topics during training, while theirs is applied to the topics generated once training is completed.}

\clearpage

\section*{Limitations}
We address several limitations with regard to our work. First, the publicly available datasets used in our experiments are limited to English. Documents in different languages (i.e., Chinese) might require different segmentation techniques and may contain unique characteristics in terms of vocabulary size, data sparsity, and ambiguity. Secondly, we only evaluate the quality of the topic models in terms of coherence and diversity. Future work should explore how our method impacts other characteristics, such as document coverage (i.e., how well documents match their assigned topics) and topic model comprehensiveness (i.e., how thoroughly the model covers the topics appearing in the corpus).  



\section*{Ethics Statement}

The datasets used in this work are publicly available and selected from recent literature. There could exist biased views in their content, and should be viewed with discretion. 

Our proposed method can be applied to extract topics from a large collection of documents. Researchers wishing to apply our method should ensure that the input corpora are adequately collected and do not violate any copyright infringements.

%




\bibliography{custom, anthology}

\begin{thebibliography}{40}
\expandafter\ifx\csname natexlab\endcsname\relax\def\natexlab#1{#1}\fi

\bibitem[{Bianchi et~al.(2021{\natexlab{a}})Bianchi, Terragni, and
  Hovy}]{bianchi-etal-2021-pre}
Federico Bianchi, Silvia Terragni, and Dirk Hovy. 2021{\natexlab{a}}.
\newblock \href {https://doi.org/10.18653/v1/2021.acl-short.96} {Pre-training
  is a hot topic: Contextualized document embeddings improve topic coherence}.
\newblock In \emph{Proceedings of the 59th Annual Meeting of the Association
  for Computational Linguistics and the 11th International Joint Conference on
  Natural Language Processing (Volume 2: Short Papers)}, pages 759--766,
  Online. Association for Computational Linguistics.

\bibitem[{Bianchi et~al.(2021{\natexlab{b}})Bianchi, Terragni, Hovy, Nozza, and
  Fersini}]{bianchi-etal-2021-cross}
Federico Bianchi, Silvia Terragni, Dirk Hovy, Debora Nozza, and Elisabetta
  Fersini. 2021{\natexlab{b}}.
\newblock \href {https://doi.org/10.18653/v1/2021.eacl-main.143} {Cross-lingual
  contextualized topic models with zero-shot learning}.
\newblock In \emph{Proceedings of the 16th Conference of the European Chapter
  of the Association for Computational Linguistics: Main Volume}, pages
  1676--1683, Online. Association for Computational Linguistics.

\bibitem[{Blei et~al.(2003)Blei, Ng, and Jordan}]{blei2003latent}
David~M Blei, Andrew~Y Ng, and Michael~I Jordan. 2003.
\newblock \href {https://jmlr.org/papers/v3/blei03a.html} {Latent dirichlet
  allocation}.
\newblock \emph{Journal of Machine Learning Research}, 3(Jan):993--1022.

\bibitem[{Bouma(2009)}]{bouma2009normalized}
Gerlof Bouma. 2009.
\newblock \href
  {https://svn.spraakdata.gu.se/repos/gerlof/pub/www/Docs/npmi-pfd.pdf}
  {Normalized (pointwise) mutual information in collocation extraction}.
\newblock \emph{Proceedings of GSCL}, 30:31--40.

\bibitem[{Burkhardt and Kramer(2019)}]{burkhardt2019decoupling}
Sophie Burkhardt and Stefan Kramer. 2019.
\newblock \href {http://jmlr.org/papers/v20/18-569.html} {Decoupling sparsity
  and smoothness in the dirichlet variational autoencoder topic model}.
\newblock \emph{Journal of Machine Learning Research}, 20(131):1--27.

\bibitem[{Card et~al.(2018)Card, Tan, and Smith}]{card-etal-2018-neural}
Dallas Card, Chenhao Tan, and Noah~A. Smith. 2018.
\newblock \href {https://doi.org/10.18653/v1/P18-1189} {Neural models for
  documents with metadata}.
\newblock In \emph{Proceedings of the 56th Annual Meeting of the Association
  for Computational Linguistics (Volume 1: Long Papers)}, pages 2031--2040,
  Melbourne, Australia. Association for Computational Linguistics.

\bibitem[{Chowdhery et~al.(2022)Chowdhery, Narang, Devlin, Bosma, Mishra,
  Roberts, Barham, Chung, Sutton, Gehrmann et~al.}]{chowdhery2022palm}
Aakanksha Chowdhery, Sharan Narang, Jacob Devlin, Maarten Bosma, Gaurav Mishra,
  Adam Roberts, Paul Barham, Hyung~Won Chung, Charles Sutton, Sebastian
  Gehrmann, et~al. 2022.
\newblock \href {https://arxiv.org/abs/2204.02311} {Palm: Scaling language
  modeling with pathways}.
\newblock \emph{arXiv preprint arXiv:2204.02311}.

\bibitem[{Dieng et~al.(2020)Dieng, Ruiz, and Blei}]{dieng-etal-2020-topic}
Adji~B. Dieng, Francisco J.~R. Ruiz, and David~M. Blei. 2020.
\newblock \href {https://doi.org/10.1162/tacl_a_00325} {Topic modeling in
  embedding spaces}.
\newblock \emph{Transactions of the Association for Computational Linguistics},
  8:439--453.

\bibitem[{Ding et~al.(2018)Ding, Nallapati, and
  Xiang}]{ding-etal-2018-coherence}
Ran Ding, Ramesh Nallapati, and Bing Xiang. 2018.
\newblock \href {https://doi.org/10.18653/v1/D18-1096} {Coherence-aware neural
  topic modeling}.
\newblock In \emph{Proceedings of the 2018 Conference on Empirical Methods in
  Natural Language Processing}, pages 830--836, Brussels, Belgium. Association
  for Computational Linguistics.

\bibitem[{Fang et~al.(2016)Fang, Macdonald, Ounis, and Habel}]{fang2016using}
Anjie Fang, Craig Macdonald, Iadh Ounis, and Philip Habel. 2016.
\newblock \href {https://doi.org/10.1145/2911451.2914729} {Using word embedding
  to evaluate the coherence of topics from twitter data}.
\newblock In \emph{Proceedings of the 39th International ACM SIGIR Conference
  on Research and Development in Information Retrieval}, SIGIR '16, page
  1057–1060, New York, NY, USA. Association for Computing Machinery.

\bibitem[{Hoffman et~al.(2010)Hoffman, Bach, and Blei}]{hoffman2010online}
Matthew Hoffman, Francis Bach, and David Blei. 2010.
\newblock \href
  {https://proceedings.neurips.cc/paper_files/paper/2010/file/71f6278d140af599e06ad9bf1ba03cb0-Paper.pdf}
  {Online learning for latent dirichlet allocation}.
\newblock In \emph{Advances in Neural Information Processing Systems},
  volume~23. Curran Associates, Inc.

\bibitem[{Hoyle et~al.(2020)Hoyle, Goel, and
  Resnik}]{hoyle-etal-2020-improving}
Alexander~Miserlis Hoyle, Pranav Goel, and Philip Resnik. 2020.
\newblock \href {https://doi.org/10.18653/v1/2020.emnlp-main.137} {{I}mproving
  {N}eural {T}opic {M}odels using {K}nowledge {D}istillation}.
\newblock In \emph{Proceedings of the 2020 Conference on Empirical Methods in
  Natural Language Processing (EMNLP)}, pages 1752--1771, Online. Association
  for Computational Linguistics.

\bibitem[{Joty et~al.(2013)Joty, Carenini, and Ng}]{joty2013topic}
Shafiq Joty, Giuseppe Carenini, and Raymond~T Ng. 2013.
\newblock \href {https://doi.org/10.1613/jair.3940} {Topic segmentation and
  labeling in asynchronous conversations}.
\newblock \emph{Journal of Artificial Intelligence Research}, 47:521--573.

\bibitem[{Kingma and Ba(2015)}]{kingma2014adam}
Diederik~P. Kingma and Jimmy Ba. 2015.
\newblock \href {http://arxiv.org/abs/1412.6980} {Adam: {A} method for
  stochastic optimization}.
\newblock In \emph{3rd International Conference on Learning Representations,
  {ICLR} 2015, San Diego, CA, USA, May 7-9, 2015, Conference Track
  Proceedings}.

\bibitem[{Kingma and Welling(2014)}]{kingma2013auto}
Diederik~P. Kingma and Max Welling. 2014.
\newblock \href {http://arxiv.org/abs/1312.6114} {Auto-encoding variational
  bayes}.
\newblock In \emph{2nd International Conference on Learning Representations,
  {ICLR} 2014, Banff, AB, Canada, April 14-16, 2014, Conference Track
  Proceedings}.

\bibitem[{Lau et~al.(2014)Lau, Newman, and Baldwin}]{lau-etal-2014-machine}
Jey~Han Lau, David Newman, and Timothy Baldwin. 2014.
\newblock \href {https://doi.org/10.3115/v1/E14-1056} {Machine reading tea
  leaves: Automatically evaluating topic coherence and topic model quality}.
\newblock In \emph{Proceedings of the 14th Conference of the {E}uropean Chapter
  of the Association for Computational Linguistics}, pages 530--539,
  Gothenburg, Sweden. Association for Computational Linguistics.

\bibitem[{Lim and Lauw(2022)}]{lim-lauw-2022-towards}
Jia~Peng Lim and Hady Lauw. 2022.
\newblock \href {https://aclanthology.org/2022.emnlp-main.242} {Towards
  reinterpreting neural topic models via composite activations}.
\newblock In \emph{Proceedings of the 2022 Conference on Empirical Methods in
  Natural Language Processing}, pages 3688--3703, Abu Dhabi, United Arab
  Emirates. Association for Computational Linguistics.

\bibitem[{Miao et~al.(2017)Miao, Grefenstette, and
  Blunsom}]{miao2017discovering}
Yishu Miao, Edward Grefenstette, and Phil Blunsom. 2017.
\newblock \href {https://proceedings.mlr.press/v70/miao17a.html} {Discovering
  discrete latent topics with neural variational inference}.
\newblock In \emph{Proceedings of the 34th International Conference on Machine
  Learning}, volume~70 of \emph{Proceedings of Machine Learning Research},
  pages 2410--2419. PMLR.

\bibitem[{Miao et~al.(2016)Miao, Yu, and Blunsom}]{miao2016neural}
Yishu Miao, Lei Yu, and Phil Blunsom. 2016.
\newblock \href {https://proceedings.mlr.press/v48/miao16.html} {Neural
  variational inference for text processing}.
\newblock In \emph{Proceedings of The 33rd International Conference on Machine
  Learning}, volume~48 of \emph{Proceedings of Machine Learning Research},
  pages 1727--1736, New York, New York, USA. PMLR.

\bibitem[{Mihalcea and Radev(2011)}]{mihalcea2011graph}
Rada Mihalcea and Dragomir Radev. 2011.
\newblock \href {https://doi.org/10.1017/CBO9780511976247} {\emph{Graph-based
  Natural Language Processing and Information Retrieval}}.
\newblock Cambridge University Press.

\bibitem[{Mimno et~al.(2011)Mimno, Wallach, Talley, Leenders, and
  McCallum}]{mimno-etal-2011-optimizing}
David Mimno, Hanna Wallach, Edmund Talley, Miriam Leenders, and Andrew
  McCallum. 2011.
\newblock \href {https://aclanthology.org/D11-1024} {Optimizing semantic
  coherence in topic models}.
\newblock In \emph{Proceedings of the 2011 Conference on Empirical Methods in
  Natural Language Processing}, pages 262--272, Edinburgh, Scotland, UK.
  Association for Computational Linguistics.

\bibitem[{Nan et~al.(2019)Nan, Ding, Nallapati, and
  Xiang}]{nan-etal-2019-topic}
Feng Nan, Ran Ding, Ramesh Nallapati, and Bing Xiang. 2019.
\newblock \href {https://doi.org/10.18653/v1/P19-1640} {Topic modeling with
  {W}asserstein autoencoders}.
\newblock In \emph{Proceedings of the 57th Annual Meeting of the Association
  for Computational Linguistics}, pages 6345--6381, Florence, Italy.
  Association for Computational Linguistics.

\bibitem[{Nevezhin et~al.(2020)Nevezhin, Butakov, Khodorchenko, Petrov, and
  Nasonov}]{nevezhin-etal-2020-topic}
Egor Nevezhin, Nikolay Butakov, Maria Khodorchenko, Maxim Petrov, and Denis
  Nasonov. 2020.
\newblock \href {https://doi.org/10.18653/v1/2020.coling-main.206}
  {Topic-driven ensemble for online advertising generation}.
\newblock In \emph{Proceedings of the 28th International Conference on
  Computational Linguistics}, pages 2273--2283, Barcelona, Spain (Online).
  International Committee on Computational Linguistics.

\bibitem[{Newman et~al.(2010)Newman, Lau, Grieser, and
  Baldwin}]{newman-etal-2010-automatic}
David Newman, Jey~Han Lau, Karl Grieser, and Timothy Baldwin. 2010.
\newblock \href {https://aclanthology.org/N10-1012} {Automatic evaluation of
  topic coherence}.
\newblock In \emph{Human Language Technologies: The 2010 Annual Conference of
  the North {A}merican Chapter of the Association for Computational
  Linguistics}, pages 100--108, Los Angeles, California. Association for
  Computational Linguistics.

\bibitem[{Qiang et~al.(2022)Qiang, Qian, Li, Yuan, and Wu}]{qiang2020short}
Jipeng Qiang, Zhenyu Qian, Yun Li, Yunhao Yuan, and Xindong Wu. 2022.
\newblock \href {https://doi.org/10.1109/TKDE.2020.2992485} {Short text topic
  modeling techniques, applications, and performance: A survey}.
\newblock \emph{IEEE Transactions on Knowledge and Data Engineering},
  34(3):1427--1445.

\bibitem[{Reimers and Gurevych(2019)}]{reimers-gurevych-2019-sentence}
Nils Reimers and Iryna Gurevych. 2019.
\newblock \href {https://doi.org/10.18653/v1/D19-1410} {Sentence-{BERT}:
  Sentence embeddings using {S}iamese {BERT}-networks}.
\newblock In \emph{Proceedings of the 2019 Conference on Empirical Methods in
  Natural Language Processing and the 9th International Joint Conference on
  Natural Language Processing (EMNLP-IJCNLP)}, pages 3982--3992, Hong Kong,
  China. Association for Computational Linguistics.

\bibitem[{Rezende et~al.(2014)Rezende, Mohamed, and
  Wierstra}]{rezende2014stochastic}
Danilo~Jimenez Rezende, Shakir Mohamed, and Daan Wierstra. 2014.
\newblock \href {https://proceedings.mlr.press/v32/rezende14.html} {Stochastic
  backpropagation and approximate inference in deep generative models}.
\newblock In \emph{Proceedings of the 31st International Conference on Machine
  Learning}, volume~32 of \emph{Proceedings of Machine Learning Research},
  pages 1278--1286, Bejing, China. PMLR.

\bibitem[{R\"{o}der et~al.(2015)R\"{o}der, Both, and
  Hinneburg}]{roder2015exploring}
Michael R\"{o}der, Andreas Both, and Alexander Hinneburg. 2015.
\newblock \href {https://doi.org/10.1145/2684822.2685324} {Exploring the space
  of topic coherence measures}.
\newblock In \emph{Proceedings of the Eighth ACM International Conference on
  Web Search and Data Mining}, WSDM '15, page 399–408, New York, NY, USA.
  Association for Computing Machinery.

\bibitem[{Srivastava and Sutton(2017)}]{srivastava2017autoencoding}
Akash Srivastava and Charles Sutton. 2017.
\newblock \href {https://openreview.net/forum?id=BybtVK9lg} {Autoencoding
  variational inference for topic models}.
\newblock In \emph{5th International Conference on Learning Representations,
  {ICLR} 2017, Toulon, France, April 24-26, 2017, Conference Track
  Proceedings}.

\bibitem[{Srivastava et~al.(2014)Srivastava, Hinton, Krizhevsky, Sutskever, and
  Salakhutdinov}]{srivastava2014dropout}
Nitish Srivastava, Geoffrey Hinton, Alex Krizhevsky, Ilya Sutskever, and Ruslan
  Salakhutdinov. 2014.
\newblock \href {http://jmlr.org/papers/v15/srivastava14a.html} {Dropout: A
  simple way to prevent neural networks from overfitting}.
\newblock \emph{Journal of Machine Learning Research}, 15(56):1929--1958.

\bibitem[{Steyvers and Griffiths(2007)}]{steyvers2007probabilistic}
Mark Steyvers and Tom Griffiths. 2007.
\newblock \href {https://doi.org/10.4324/9780203936399} {Probabilistic topic
  models}.
\newblock In \emph{Handbook of Latent Semantic Analysis}, pages 439--460.
  Psychology Press.

\bibitem[{Terragni et~al.(2021)Terragni, Fersini, and
  Messina}]{terragni2021word}
Silvia Terragni, Elisabetta Fersini, and Enza Messina. 2021.
\newblock \href {https://doi.org/10.1007/978-3-030-80599-9_4} {Word
  embedding-based topic similarity measures}.
\newblock In \emph{Natural Language Processing and Information Systems: 26th
  International Conference on Applications of Natural Language to Information
  Systems, NLDB 2021, Saarbr{\"u}cken, Germany, June 23--25, 2021,
  Proceedings}, pages 33--45. Springer.

\bibitem[{Wang et~al.(2019)Wang, Gan, Xu, Zhang, Wang, Shen, Chen, and
  Carin}]{wang-etal-2019-topic}
Wenlin Wang, Zhe Gan, Hongteng Xu, Ruiyi Zhang, Guoyin Wang, Dinghan Shen,
  Changyou Chen, and Lawrence Carin. 2019.
\newblock \href {https://doi.org/10.18653/v1/N19-1015} {Topic-guided
  variational auto-encoder for text generation}.
\newblock In \emph{Proceedings of the 2019 Conference of the North {A}merican
  Chapter of the Association for Computational Linguistics: Human Language
  Technologies, Volume 1 (Long and Short Papers)}, pages 166--177, Minneapolis,
  Minnesota. Association for Computational Linguistics.

\bibitem[{Wang et~al.(2020)Wang, Duan, Zhang, Wang, Tian, Chen, and
  Zhou}]{wang-etal-2020-friendly}
Zhengjue Wang, Zhibin Duan, Hao Zhang, Chaojie Wang, Long Tian, Bo~Chen, and
  Mingyuan Zhou. 2020.
\newblock \href {https://doi.org/10.18653/v1/2020.emnlp-main.35} {Friendly
  topic assistant for transformer based abstractive summarization}.
\newblock In \emph{Proceedings of the 2020 Conference on Empirical Methods in
  Natural Language Processing (EMNLP)}, pages 485--497, Online. Association for
  Computational Linguistics.

\bibitem[{Xiao et~al.(2022)Xiao, Miculicich, Liu, He, and
  Carenini}]{xiao2022attend}
Wen Xiao, Lesly Miculicich, Yang Liu, Pengcheng He, and Giuseppe Carenini.
  2022.
\newblock \href {https://arxiv.org/abs/2212.10819} {Attend to the right
  context: A plug-and-play module for content-controllable summarization}.
\newblock \emph{arXiv preprint arXiv:2212.10819}.

\bibitem[{Xing et~al.(2019)Xing, Paul, and
  Carenini}]{xing-etal-2019-evaluating}
Linzi Xing, Michael~J. Paul, and Giuseppe Carenini. 2019.
\newblock \href {https://doi.org/10.18653/v1/D19-1349} {Evaluating topic
  quality with posterior variability}.
\newblock In \emph{Proceedings of the 2019 Conference on Empirical Methods in
  Natural Language Processing and the 9th International Joint Conference on
  Natural Language Processing (EMNLP-IJCNLP)}, pages 3471--3477, Hong Kong,
  China. Association for Computational Linguistics.

\bibitem[{Xu et~al.(2021)Xu, Zhao, and Zhang}]{xu2021topic}
Yi~Xu, Hai Zhao, and Zhuosheng Zhang. 2021.
\newblock \href {https://doi.org/10.1609/aaai.v35i16.17668} {Topic-aware
  multi-turn dialogue modeling}.
\newblock \emph{Proceedings of the AAAI Conference on Artificial Intelligence},
  35(16):14176--14184.

\bibitem[{Zhang et~al.(2022)Zhang, Hu, Wang, Zhou, Zhang, and
  Cao}]{zhang-etal-2022-pre}
Linhai Zhang, Xuemeng Hu, Boyu Wang, Deyu Zhou, Qian-Wen Zhang, and Yunbo Cao.
  2022.
\newblock \href {https://doi.org/10.18653/v1/2022.acl-long.413} {Pre-training
  and fine-tuning neural topic model: A simple yet effective approach to
  incorporating external knowledge}.
\newblock In \emph{Proceedings of the 60th Annual Meeting of the Association
  for Computational Linguistics (Volume 1: Long Papers)}, pages 5980--5989,
  Dublin, Ireland. Association for Computational Linguistics.

\bibitem[{Zheng et~al.(2015)Zheng, Yang, Liu, Liang, and
  Li}]{zheng2015improving}
Hao Zheng, Zhanlei Yang, Wenju Liu, Jizhong Liang, and Yanpeng Li. 2015.
\newblock \href {https://doi.org/10.1109/IJCNN.2015.7280459} {Improving deep
  neural networks using softplus units}.
\newblock In \emph{2015 International Joint Conference on Neural Networks
  (IJCNN)}, pages 1--4. IEEE.

\bibitem[{Zhu et~al.(2021)Zhu, Pergola, Gui, Zhou, and
  He}]{zhu-etal-2021-topic}
Lixing Zhu, Gabriele Pergola, Lin Gui, Deyu Zhou, and Yulan He. 2021.
\newblock \href {https://doi.org/10.18653/v1/2021.acl-long.125} {Topic-driven
  and knowledge-aware transformer for dialogue emotion detection}.
\newblock In \emph{Proceedings of the 59th Annual Meeting of the Association
  for Computational Linguistics and the 11th International Joint Conference on
  Natural Language Processing (Volume 1: Long Papers)}, pages 1571--1582,
  Online. Association for Computational Linguistics.

\end{thebibliography}
\bibliographystyle{acl_natbib}

\clearpage
\appendix

\section{LDA Generative Process}
\label{sec:lda}
The formal generative process of a corpus under the LDA assumption can be described by the following algorithm.
\begin{algorithm}[ht!]
\caption{Generative process of LDA}
\begin{algorithmic}
\For{each document $w$ \do}
     \State Sample topic distribution $\theta \sim \textrm{Dirichlet}(\alpha)$
     \For{each word $w_i$ \do}
     \State Sample topic $z_i \sim \textrm{Multinomial}(\theta)$
     \State Sample word $w_i \sim \textrm{Multinomial}(\beta_{z_i})$
    \EndFor
\EndFor
\end{algorithmic}
\end{algorithm}

\section{Normalized Pointwise Mutual Information}
\label{sec:npmi}
{Normalized Pointwise Mutual Information (NPMI)~\citep{lau-etal-2014-machine} 
measures how much more likely the most representative terms of a topic co-occur than if they were independent. The method for computing the NPMI score between word $w_i$ and $w_j$ is described in \autoref{eq:npmi}, where $P(w_i,w_j)$ is computed using a window size of $10$. This metric ranges from $-1$ to $1$.

\begin{equation}
\label{eq:npmi}
   \textrm{NPMI}(w_i, w_j) =  \frac{\log\frac{P(w_{i}, w_j)}{P(w_{i})P(w_{j})}}{-\log P(w_{i},w_{j})}
\end{equation}

\change{In practice, the pairwise NPMI matrix is computed by first counting the word co-occurrence of all words in the corpus and then calculating the pairwise score following \autoref{eq:npmi}. In summary, the NPMI matrix can be computed in $\mathcal{O}(|W| + |V|^2)$ for a corpus of $|W|$ words and vocab size $|V|$. Since the matrix is  computed only once for each corpus prior to training, it does not increase the runtime complexity of training time.}
\newpage

\section{Datasets}
\label{sec:datasets}
This section provides details regarding the datasets we used. The 20NewsGroup\footnote{\url{http://qwone.com/~jason/20Newsgroups}} dataset is a collection of email documents partitioned evenly across 20 categories~(e.g., electronics, space), we use the same filtered subset provided by \citet{bianchi-etal-2021-pre}.
The Wiki20K dataset\footnote{\url{https://github.com/vinid/data}}
contains randomly sampled subsets from the English Wikipedia abstracts from DBpedia\footnote{\url{https://wiki.dbpedia.org/downloads-2016-10}}. GoogleNews\footnote{\url{https://github.com/qiang2100/STTM/tree/master/dataset}}~\citep{qiang2020short} is
downloaded from the Google news site by crawling the titles and snippets. We do not perform any additional pre-processing and directly use the data provided by the sources to create contextualized and BoW representation. 

\section{Sample Output}
\label{sec:model-output}
\autoref{tab:examples_topics_20ng} provides a qualitative comparison of the topics generated by our proposed method using ZeroshotTM on the 20NewsGroups dataset.

\section{Implementation Details}
\label{implementation}
We base our implementation using the code provided by the authors of ZeroshotTM and CombinedTM \citep{bianchi-etal-2021-pre, bianchi-etal-2021-cross}.
Their repository\footnote{\url{https://github.com/MilaNLProc/contextualized-topic-models}} also provides the evaluation metrics used in our experiments. Our Python code base includes external open-source libraries including NumPy\footnote{\url{https://numpy.org/}}, 
SciPy\footnote{\url{https://scipy.org/}},
PyTorch\footnote{\url{https://pytorch.org/}},
SentenceTransformers\footnote{\url{https://www.sbert.net/}},
Pandas\footnote{\url{https://pandas.pydata.org/}}, Gensim\footnote{\url{https://radimrehurek.com/gensim/}} and scikit-learn\footnote{\url{https://scikit-learn.org/stable/}}.

\section{Computing Details}
\label{sec:computing-details}
All our experiments are run on Linux machines with single 1080Ti GPU (CUDA version $11.4$). Each epoch with $100$ batch size on the most computationally intensive setting (GoogleNews with $K = 150$) takes on average $3$ seconds to run for the baselines models and $8$ and $15$ seconds, for $W_C$ and $W_D$, respectively. Under this setting, a maximum VRAM usage of 800MB was recorded.

\clearpage
\onecolumn
\begin{scriptsize}
\begin{longtable}{ll}
\caption{Sample model output $K=25$ by running ZeroshotTM (Z) with our proposed method ($+W_C$ and $+W_D$) on the 20NewsGroups dataset. We visualize the top-10 keywords of each topic with unique keywords in \textbf{bold}.}
\label{tab:examples_topics_20ng} \\
\toprule
 Model &  Top-10 Topic Keywords\\ \midrule
Z & newsletter, aids, hiv, medical, cancer, disease, page, health, volume, patients  \\
 Z + $W_C$ &  newsletter, aids, hiv, medical, cancer, disease, page, health, volume, patients\\
 Z + $W_D$&  newsletter, hiv, aids, medical, cancer, disease, health, page, volume, patients\\ \midrule
 Z & mary, sin, god, heaven, lord, christ, jesus, grace, spirit, \textbf{matthew}  \\
 Z + $W_C$ & mary, sin, heaven, god, christ, lord, jesus, spirit, grace, \textbf{matthew} \\
 Z + $W_D$&  mary, heaven, sin, christ, god, spirit, lord, jesus, \textbf{holy}, grace  \\
 \midrule
 Z & engine, car, bike, cars, oil, ride, road, dealer, \textbf{miles}, riding  \\
 Z + $W_C$ & engine, bike, car, cars, oil, ride, dealer, road, riding, \textbf{driving}  \\
 Z + $W_D$& engine, bike, car, cars, oil, ride, dealer, riding, road, \textbf{driving}   \\
 \midrule
   Z & game, baseball, ball, season, fans, team, year, playing, players, \textbf{winning}  \\
 Z + $W_C$ &game, baseball, fans, ball, season, team, playing, \textbf{teams}, players, year  \\
 Z + $W_D$&   baseball, game, fans, season, \textbf{teams}, ball, team, playing, players, year  \\
  \midrule
   Z & fbi, koresh, batf, trial, compound, gas, investigation, \textbf{media}, branch, agents \\
 Z + $W_C$ & fbi, batf, koresh, compound, gas, agents, trial, branch, investigation, \textbf{waco}  \\
 Z + $W_D$ & fbi, koresh, batf, compound, gas, agents, trial, branch, \textbf{waco}, investigation \\
   \midrule
   Z &  entry, rules, entries, email, build, info, file, char, program, section \\
 Z + $W_C$ & entry, rules, entries, email, info, build, file, char, section, program   \\
 Z + $W_D$& entry, rules, entries, email, build, info, file, char, program, section \\
   \midrule
   Z & army, turkey, muslim, jews, greek, jewish, genocide, \textbf{professor}, ottoman, greece  \\
 Z + $W_C$ & army, muslim, turkey, ottoman, jews, greek, genocide, jewish, greece, \textbf{muslims} \\
 Z + $W_D$&  muslim, turkey, ottoman, genocide, army, jews, greek, jewish, greece, \textbf{muslims}\\
   \midrule
   Z & board, driver, video, cards, card, monitor, windows, drivers, screen, \textbf{resolution}  \\
 Z + $W_C$ &   board, video, driver, cards, monitor, card, windows, drivers, screen, \textbf{printer}\\
 Z + $W_D$&  video, board, driver, cards, monitor, card, drivers, \textbf{printer}, screen, windows\\
   \midrule
   Z &   frequently, previously, suggested, \textbf{announced}, \textbf{foundation}, \textbf{spent}, \textbf{contain}, \textbf{grant}, \textbf{consistent}, authors\\
 Z + $W_C$ & \textbf{basically}, previously, frequently, \textbf{generally}, suggested, \textbf{primary}, authors, \textbf{appropriate}, \textbf{kinds}, \textbf{greater}  \\
 Z + $W_D$& \textbf{essentially}, \textbf{basically}, \textbf{kinds}, \textbf{consistent}, frequently, authors, previously, \textbf{primary}, \textbf{equivalent}, suggested \\
   \midrule
   Z & sale, condition, offer, asking, offers, shipping, items, price, \textbf{email}, sell  \\
 Z + $W_C$ &sale, condition, offer, shipping, asking, items, offers, sell, \textbf{email}, price \\
 Z + $W_D$& sale, condition, shipping, offer, asking, items, offers, sell, price, \textbf{excellent} \\
   \midrule
   Z & application, window, xterm, motif, font, manager, widget, \textbf{root}, event, server  \\
 Z + $W_C$ & xterm, application, window, motif, font, widget, manager, \textbf{x11r5}, server, event  \\
 Z + $W_D$& xterm, motif, font, application, window, widget, manager, \textbf{x11r5}, event, server \\
 \midrule
   Z & gun, amendment, constitution, firearms, right, militia, guns, weapon, bear, weapons \\
 Z + $W_C$ & amendment, constitution, firearms, gun, militia, right, guns, weapon, bear, weapons \\
 Z + $W_D$& amendment, firearms, constitution, gun, militia, guns, right, weapon, bear, weapons\\
 \midrule
   Z & \textbf{suggested}, \textbf{frequently}, previously, \textbf{authors}, \textbf{foundation}, \textbf{consistent}, \textbf{spent}, \textbf{join}, \textbf{et}, \textbf{announced} \\
 Z + $W_C$ & \textbf{suggested}, previously, \textbf{frequently}, \textbf{greater}, \textbf{requirements}, \textbf{consistent}, \textbf{opportunity}, \textbf{authors}, \textbf{particularly}, \textbf{appropriate} \\
 Z + $W_D$& \textbf{spent}, \textbf{greater}, \textbf{association}, \textbf{appropriate}, \textbf{opportunity}, \textbf{requirements}, \textbf{posts}, previously, \textbf{success}, \textbf{training}\\
 \midrule
   Z & objective, atheist, atheism, morality, exists, belief, does, exist, atheists, existence \\
 Z + $W_C$ & objective, atheist, atheism, morality, exists, belief, atheists, does, exist, existence \\
 Z + $W_D$& atheist, objective, atheism, belief, morality, exists, atheists, existence, exist, does\\
 \midrule
   Z &think, president, people, Stephanopoulos, dont, jobs, just, know, mr, myers  \\
 Z + $W_C$ & think, president, Stephanopoulos, people, dont, jobs, just, know mr, myers \\
 Z + $W_D$&think, president, Stephanopoulos, people, dont, jobs, just, know, mr, myers \\
 \midrule
   Z &  board, drive, ide, scsi, bus, isa, mhz, motherboard, \textbf{internal}, \textbf{pin}\\
 Z + $W_C$ & board, drive, ide, scsi, motherboard, bus, isa, mhz, \textbf{hd}, \textbf{controller} \\
 Z + $W_D$&board, drive, ide, motherboard, scsi, mhz, bus, \textbf{hd}, isa, \textbf{controller} \\
 \midrule
   Z &  jpeg, images, image, formats, gif, format, software, conversion, quality, color\\
 Z + $W_C$ & jpeg, images, formats, image, gif, format, conversion, software, quality, color \\
 Z + $W_D$& jpeg, images, formats, gif, image, format, conversion, software, quality, color\\
 \midrule
   Z & msg, food, doctor, vitamin, doctors, medicine, diet, \textbf{insurance}, treatment, studies \\
 Z + $W_C$ & msg, food, doctor, medicine, doctors, vitamin, diet, studies, treatment, \textbf{insurance} \\
 Z + $W_D$& msg, food, doctor, medicine, doctors, vitamin, diet, studies, \textbf{patients}, treatment\\
 \midrule
   Z & agencies, encryption, keys, secure, algorithm, \textbf{chip}, enforcement, nsa, \textbf{clipper}, \textbf{secret} \\
 Z + $W_C$ &  agencies, encryption, secure, keys, algorithm, nsa, enforcement, \textbf{encrypted}, \textbf{escrow}, \textbf{chip}\\
 Z + $W_D$& secure, encryption, keys, agencies, algorithm, \textbf{escrow}, \textbf{encrypted}, enforcement, nsa, \textbf{clipper}\\
 \midrule
   Z &  windows, dos, nt, network, card, disk, pc, software, modem, operating\\
 Z + $W_C$ &  windows, dos, nt, card, network, disk, pc, modem, software, operating\\
 Z + $W_D$& windows, dos, nt, card, network, disk, pc, modem, software, operating\\
 \midrule
   Z & address, site, thanks, looking, newsgroup, appreciate, advance, mailing, \textbf{obtain}, \textbf{domain} \\
 Z + $W_C$ &address, thanks, newsgroup, site, appreciate, advance, looking, mailing, \textbf{thank}, \textbf{reply}  \\
 Z + $W_D$&address, appreciate, site, thanks, advance, newsgroup, looking, mailing, \textbf{thank}, \textbf{obtain} \\
 \bottomrule \\\\\\\\
 \toprule
 Model &  Top-10 Topic Keywords\\ \midrule
   Z &launch, nasa, shuttle, mission, satellite, energy, mass, moon, orbit, lunar  \\
 Z + $W_C$ & launch, shuttle, nasa, mission, moon, satellite, orbit, energy, mass, lunar \\
 Z + $W_D$& shuttle, launch, nasa, mission, orbit, moon, satellite, lunar, mass, energy\\
 \midrule
   Z &  floor, door, said, people, azerbaijani, neighbors, apartment, like, saw, \textbf{dont}\\
 Z + $W_C$ & floor, azerbaijani, door, said, people, apartment, neighbors, like, saw, \textbf{dont} \\
 Z + $W_D$&azerbaijani, floor, apartment, door, said, people, neighbors, saw, like, \textbf{building}\\
 \midrule
   Z &  join, \textbf{grant}, \textbf{foundation}, \textbf{suggested}, \textbf{previously}, discussions, \textbf{frequently}, \textbf{authors}, \textbf{positions}, \textbf{announced}\\
 Z + $W_C$ & discussions, \textbf{topic}, \textbf{suggested}, join, \textbf{mailing}, \textbf{responses}, \textbf{robert}, \textbf{lists}, \textbf{summary}, \textbf{received} \\
 Z + $W_D$&join, discussions, \textbf{foundation}, \textbf{robert}, \textbf{mailing}, \textbf{lists}, \textbf{topic}, \textbf{grant}, \textbf{received}, \textbf{responses} \\
 \midrule
   Z &  pts, boston, van, pittsburgh, pp, san, \textbf{vancouver}, chicago, \textbf{la}, \textbf{st}\\
 Z + $W_C$ &  pts, boston, van, pittsburgh, pp, san, \textbf{vancouver}, chicago, \textbf{buf}, \textbf{tor}\\
 Z + $W_D$&pts, pittsburgh, van, boston, pp, chicago, \textbf{buf}, \textbf{tor}, san, \textbf{det} \\
\bottomrule
\end{longtable}
\end{scriptsize}
\clearpage
\twocolumn


\end{document}